\documentclass[10pt,twocolumn,letterpaper]{article}

\usepackage{iccv}
\usepackage{times}
\usepackage{epsfig}
\usepackage{graphicx}
\usepackage{amsmath}
\usepackage{amssymb}
\usepackage{bbm}
\usepackage{url}
\usepackage{multirow}
\usepackage{subcaption}
\usepackage{expl3}
\usepackage{enumitem}

\usepackage[pagebackref=true,breaklinks=true,letterpaper=true,colorlinks,bookmarks=false]{hyperref}

\iccvfinalcopy

\ExplSyntaxOn
\newcommand\latinabbrev[1]{
  \peek_meaning:NTF . {
    #1\@}%
  { \peek_catcode:NTF a {
      #1.\@ }%
    {#1.\@}}}
\ExplSyntaxOff
\def\etc{\latinabbrev{etc}}

\def\iccvPaperID{1737} 

\ificcvfinal\fi

\begin{document}

\title{Sparse Autoencoder for Unsupervised Nucleus Detection and Representation in Histopathology Images}

\author{
Le Hou$^1$, Vu Nguyen$^1$, Dimitris Samaras$^1$, Tahsin M. Kurc$^{1,2}$,
       Yi Gao$^{1}$, Tianhao Zhao$^1$, Joel H. Saltz$^{1,3}$
       \\
       $^1$Stony Brook University, $^2$Oak Ridge National Laboratory, $^3$Stony Brook University Hospital\\
}

\maketitle

\begin{abstract}
Histopathology images are crucial to the study of complex diseases such as cancer. The histologic characteristics of nuclei play a key role in disease diagnosis, prognosis and analysis. In this work, we propose a sparse Convolutional Autoencoder (CAE) for fully unsupervised, simultaneous nucleus detection and feature extraction in histopathology tissue images. Our CAE detects and encodes nuclei in image patches in tissue images into sparse feature maps that encode both the location and appearance of nuclei. Our CAE is the first unsupervised detection network for computer vision applications. The pretrained nucleus detection and feature extraction modules in our CAE can be fine-tuned for supervised learning in an end-to-end fashion. We evaluate our method on four datasets and reduce the errors of state-of-the-art methods up to 42\%. We are able to achieve comparable performance with only 5\% of the fully-supervised annotation cost.
\end{abstract}

\section{Introduction}
Pathologists routinely examine glass tissue slides for disease diagnosis in healthcare settings. Nuclear characteristics, such as size, shape and chromatin pattern, are important factors in distinguishing different types of cells and diagnosing disease stage. Manual examination of glass slides, however, is not feasible in large scale research studies which may involve thousands of slides.  Automatic analysis of nuclei can provide quantitative measures and new insights to disease mechanisms that cannot be gleaned from manual, qualitative evaluations of tissue specimens.  

\begin{figure}[t]
\begin{center}
   \includegraphics[trim=0cm 0.3cm 0cm 0.3cm,clip, width=0.96\linewidth]{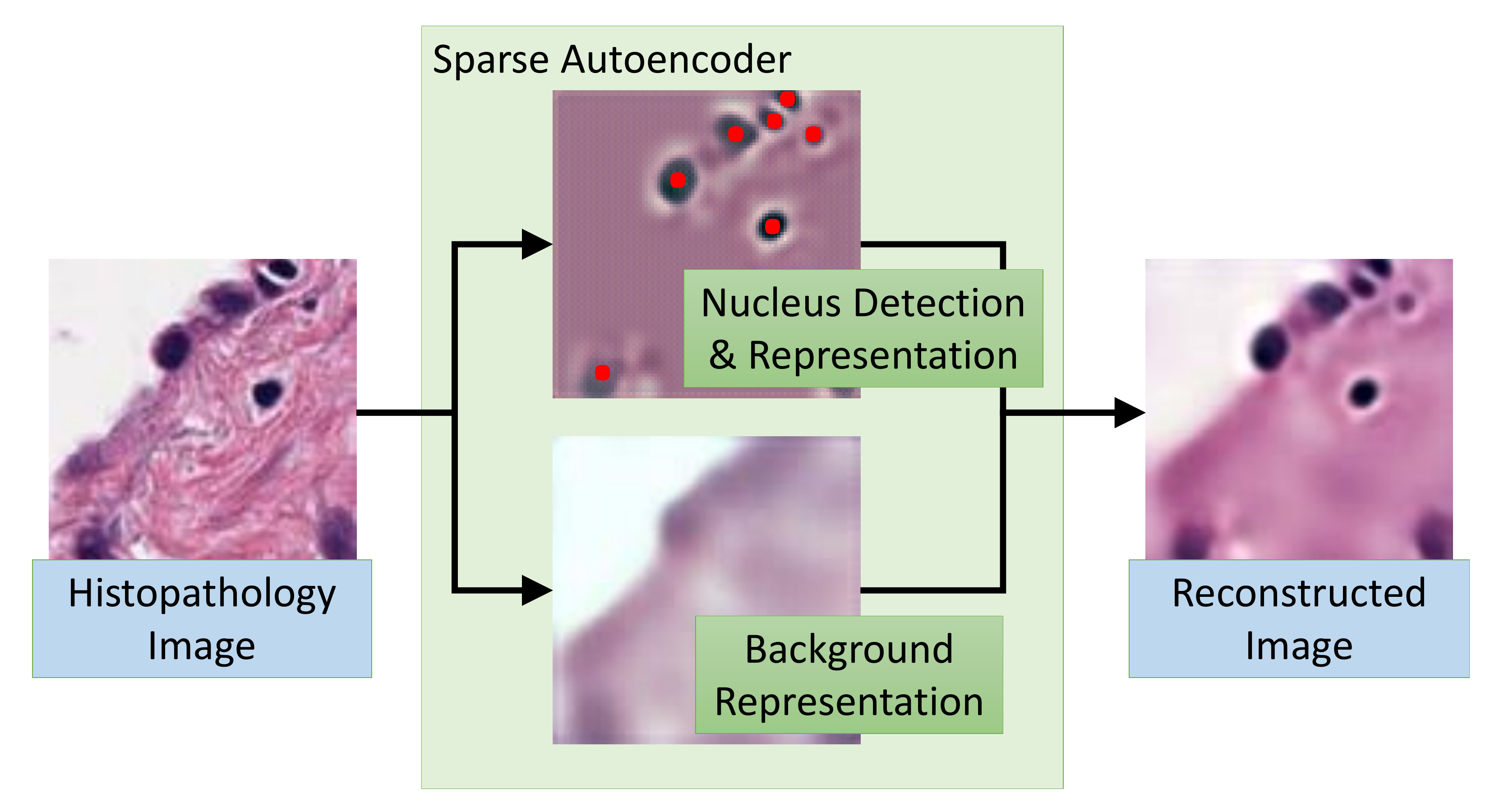}
\end{center}
\vspace{-0.25cm}
   \caption{Our fully unsupervised autoencoder. It first decomposes an input histopathology image patch into foreground (\eg nuclei) and background (\eg cytoplasm). It then detects nuclei in the foreground by representing the locations of nuclei as a sparse feature map. Finally, it encodes each nucleus to a feature vector. Our autoencoder is trained end-to-end, minimizing the reconstruction error.}
\label{fig:first_fig}
\end{figure}

Collecting a large-scale supervised dataset is a labor intensive
and challenging process since it requires the involvement
of expert pathologists who's time is a very limited and expensive resource~\cite{pathologistwages}. Thus existing state-of-the-art nucleus analysis methods are semi-supervised~\cite{LeCun:2007,LeCun:2008,masci2011stacked,bayramoglu2016transfer,xu2016stacked}: 1). Pretrained an autoencoder for unsupervised representation learning; 2). Construct a CNN from the pretrained autoencoder; 3). Fine-tune the constructed CNN for supervised nucleus classification. To better capture the visual variance of nuclei, one usually trains the unsupervised autoencoder on image patches with nuclei in the center~\cite{hou2016automatic,murthy2017center}. This requires a separate nucleus detection step~\cite{sirinukunwattana2016locality} which in most cases needs tuning to optimize the final classification performance.

\begin{figure*}[t]
\begin{center}
   \includegraphics[trim=0cm 0.1cm 0cm 0.1cm,clip, width=0.98\linewidth]{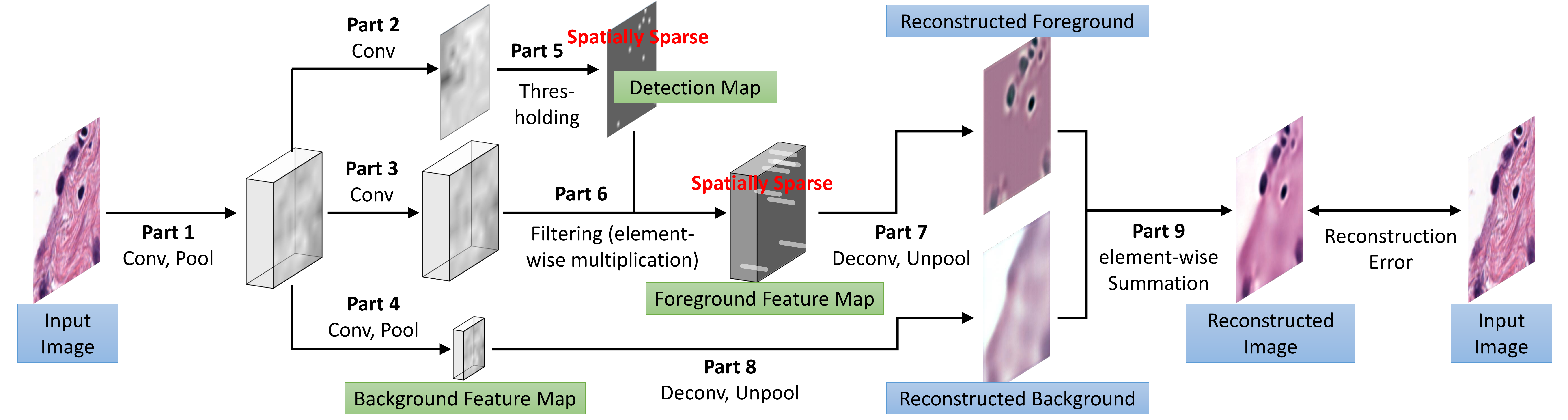}
\end{center}
\vspace{-0.25cm}
   \caption{The architecture of our sparse Convolutional Autoencoder (CAE). The CAE minimizes image reconstruction error. The reconstructed image patch is a pixel-wise summation of two intermediate reconstructed image patches: the background and the foreground. The background is reconstructed from a set of small feature maps (background feature map) that can only encode large scale color and texture. The foreground is reconstructed from a set of crosswise sparse feature maps (foreground feature map). These foreground feature maps capture local high frequency signal: nuclei. We define \textit{crosswise sparsity} as a type of sparsity with the following constraint: when there is no detected nucleus at a location, neurons in all foreground feature maps at the same location should not be activated. Details of network parts 1-8 are in Tab.~\ref{tab:net_layers}.}
\label{fig:pipeline}
\end{figure*}

Instead of tuning the detection and classification modules separately, recent works~\cite{graves2014towards,ren2015faster,redmon2016you,kokkinos2016ubernet} successfully trained end-to-end CNNs to perform these tasks in an unified pipeline. Prior work has developed and employed supervised networks. To utilize unlabeled data for unsupervised pretraining, a network that can be trained end-to-end must perform unsupervised nucleus detection. Such unsupervised detection networks do not exist in any visual application domains, despite the success of unsupervised learning in other tasks~\cite{radford2015unsupervised,doersch2015unsupervised}.


We design a novel Convolutional Autoencoder (CAE) that unifies nuclei detection and feature/representation learning in a single network and can be trained end-to-end without supervision. We also show that with existing labeled data, our network can be easily fine-tuned with supervision to improve the state-of-the-art performance of nuclei classification and segmentation.

To perform unsupervised representation learning and detection, we modify the conventional CAE to encode not only appearance, but also spatial information in feature maps. To this end, our CAE first learns to separate background (\eg cytoplasm) and foreground (\eg nuclei) in an image patch, as shown in Fig.~\ref{fig:pipeline}. We should note that an image patch is a rectangular region in a whole slide tissue image. We use image patches, because a tissue image can be very large and may not fit in memory. It is common in tissue image analysis to partition tissue images into patches and process the patches. We will refer to the partitioned image patches simply as the images. The CAE encodes the input image in a set of low resolution feature maps (background feature maps) with a small number of encoding neurons. The feature maps can only encode large scale color and texture variations because of their limited capacity. Thus these feature maps encode the image background. The high frequency residual between the input image and the reconstructed background is the foreground that contains nuclei.

The set of nuclei is often sparse in the image. We utilize this sparse property to design the foreground learning sub-network. Specifically, we design our network to learn the foreground feature maps in a ``crosswise sparse'' manner: neurons across all feature maps are not activated (output zero) in most feature map locations. Only neurons in a few feature map locations can be activated. Since the non-activated neurons have no influence in the later decoding layers, the image foreground is reconstructed using only the non-zero responses in the foreground encoding feature maps. This means that the image reconstruction error will be minimized only if the activated encoding neurons at different locations capture salient objects- the detected nuclei.

Learning a set of crosswise sparse foreground encoding feature maps is not straightforward. Neurons at the same location across all foreground feature maps should be synchronized: they should be activated or not activated at the same time depending on the presence of nuclei. In order to achieve this synchronization, the CAE needs to learn the locations of nuclei by optimizing the reconstruction error. Hence, the nucleus detection and feature extraction models are learned simultaneously during optimization. To represent the inferred nuclear locations, we introduce a special binary feature map: the nucleus detection map. We make this map sparse by thresholding neural activations. After optimization, a neuron in the nucleus detection map should output 1, if and only if there is a detected nucleus at the neuron's location. The foreground feature maps are computed by element-wise multiplications between the nucleus detection map and a set of dense feature maps (Fig.~\ref{fig:pipeline}).

In summary, our contributions are as follows.
\begin{enumerate}[leftmargin=0.2in]
\setlength\itemsep{0.18em}
\item We propose a CAE architecture with crosswise sparsity that can \textit{detect} and represent nuclei in histopathology images with the following advantages:
\begin{itemize}[leftmargin=0.1in]
\item As far as we know, this is the first unsupervised detection network for computer vision applications.
\item Our method can be fine-tuned for end-to-end supervised learning.
\end{itemize}
\item Our experimental evaluation using multiple datasets shows the proposed approach performs significantly better than other methods. The crosswise constraint in our method boosts the performance substantially. 
\begin{itemize}[leftmargin=0.1in]
\item Our method reduces the error of the best performing baseline by up to 42\% on classification tasks, and reduces the error of the U-net method~\cite{ronneberger2015u} by 20\% in nucleus segmentation.
\item Our method achieves comparable results with 5\% of training data needed by other methods, resulting in considerable savings in the cost of label generation.
\end{itemize}
\end{enumerate}

\section{Crosswise Sparse CAE}
\label{sec:our_method}
In this section we first introduce the The Convolutional Autoencoder (CAE) then describe our crosswise sparse CAE in detail. Our CAE reconstructs an image as the pixel-wise summation of a reconstructed foreground image and a reconstructed background image. The sparse foreground encoding feature maps represent detected nucleus locations and extracted nuclear features. The background feature maps are not necessarily sparse.

\subsection{CAE for Semi-supervised CNN}
\label{sec:autoencoder}
An autoencoder is an unsupervised neural network that learns to reconstruct its input. The main purpose of this model is to learn a compact representation of the input as a set of neural responses~\cite{deng2010binary}. A typical feedforward autoencoder is composed of an encoder and a decoder, which are separate layers. The encoding layer models the appearance information of the input. The decoder reconstructs the input from neural responses in the encoding layer. The CAE~\cite{masci2011stacked} and sparse CAE~\cite{LeCun:2007,LeCun:2008,xu2016stacked,ng2011sparse} are autoencoder variants. One can construct a CNN with a trained CAE. Such semi-supervised CNNs outperform fully supervised CNNs significantly in many applications~\cite{johnson2015semi,hou2016automatic}.



\subsection{Overall Architecture of Our CAE}
The architecture of our CAE is shown in Fig.~\ref{fig:pipeline}. We train the CAE to minimize the input image reconstruction error. The early stages of the CAE network consists of six convolutional and two average-pooling layers. The network then splits into three branches: the nucleus detection branch, the foreground feature branch, and the background branch. The detection branch merges into the foreground feature branch to generate the foreground feature maps that represent nuclei. The foreground and background feature maps are decoded to generate the foreground and background reconstructed images. Finally the two intermediate images are summed to form the final reconstructed image.

\subsection{Background Encoding Feature Maps}
We first model the background (tissue, cytoplasm \etc) then extract the foreground that contains nuclei. The biggest part of tissue images is background; its texture and color vary usually in a larger scale compared to the foreground. Also, usually a major portion of a tissue image is background. Thus, we encode an input image to a few small feature maps and assume those feature maps mostly contain the background information. In practice we represent the background of a $100 \times 100$ image by five $5 \times 5$ maps.

\begin{figure}[t]
\begin{center}
   \includegraphics[trim=0cm 0.1cm 0cm 0.1cm,clip, width=0.98\linewidth]{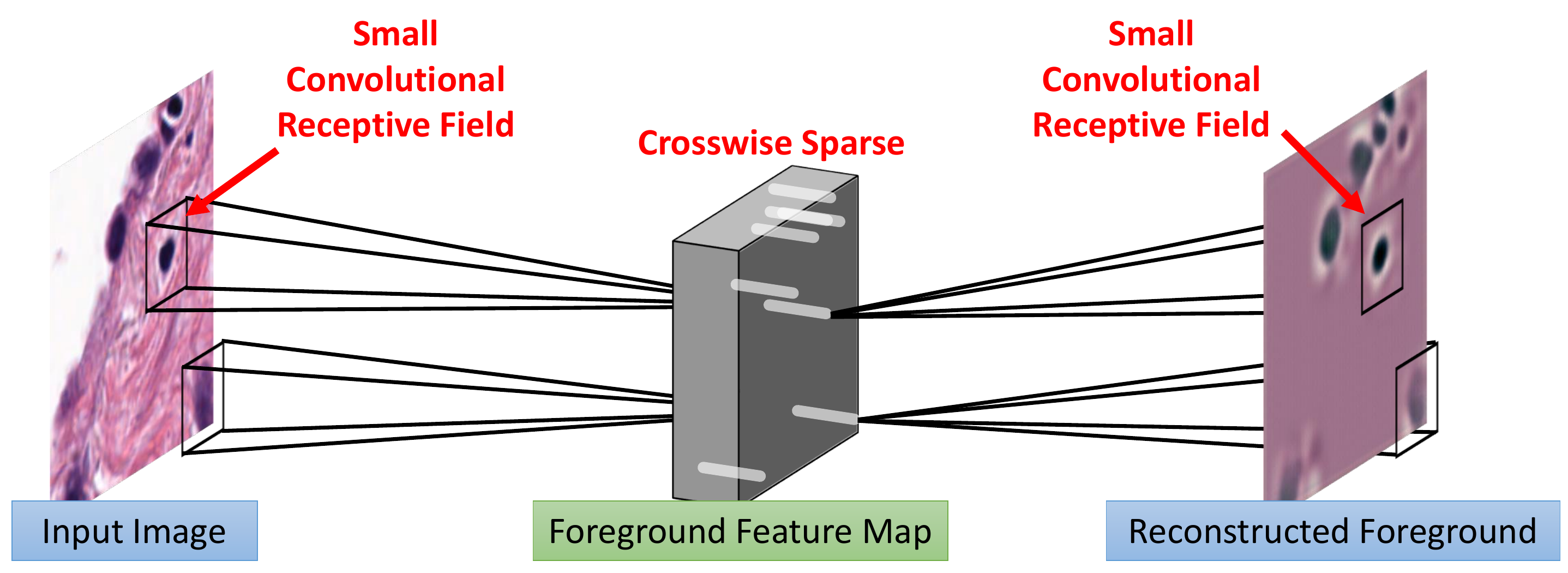}
\end{center}
\vspace{-0.25cm}
   \caption{An illustration of how each nucleus is encoded and reconstructed. First, the foreground feature map must be crosswise sparse (Eq.~\ref{eq:spartial_sparsity}). Second, the size of the receptive field of each encoding neuron should be small enough that it contains only one nucleus in most cases.}
\label{fig:foreground}
\end{figure}

\subsection{Foreground Encoding Feature Maps}
Once the background is encoded and then reconstructed, the residual between the reconstructed 
background and the input image will be the foreground. 
The foreground consists of nuclei which are roughly of the same scale and often dispersed throughout the image. The foreground encoding feature maps encode everything about the nuclei, including their locations and appearance. A foreground feature map can be viewed as a matrix, in which each entry is a vector (a set of neuron responses) that encodes an image patch (the neurons' receptive field). The vectors will encode nuclei, if there are nuclei at the center of the image patches. Otherwise the vectors contain zeros only. Since a small number of non-zero vectors encode nuclei, the foreground feature map will be sparse.

\subsubsection{Crosswise Sparsity}
\label{sec:traverse_sparsity}
We formally define crosswise sparsity as follows: We denote a set of $f$ feature maps as $X_1$, $X_2$, \dots $X_f$. Each feature map is a matrix. We denote the $i,j$-th entry of the $l$-th feature map as $X_l^{i,j}$, and the size of a feature map is $s \times s$. A conventional sparsity constraint requires:
\begin{equation}
\dfrac{\sum_{i,j,l} \mathbbm{1}(X_l^{i,j} \ne 0)}{f s^2} \ll 1 \text{,}
\label{eq:sparsity}
\end{equation}where $\mathbbm{1}(\cdot)$ is the function that returns 1 if its input is true and 0 otherwise. Crosswise sparsity requires:
\begin{equation}
\dfrac{\sum_{i,j} \mathbbm{1}\Big(\sum_l \mathbbm{1}(X_l^{i,j} \ne 0) \, > \, 0\Big)}{s^2} \ll 1 \text{.}
\label{eq:spartial_sparsity}
\end{equation}
In other words, in most locations in the foreground feature maps, neurons across \textit{all} the feature maps should \textit{not} be activated. This sparsity definition, illustrated in Fig. \ref{fig:foreground}, can be viewed as a special form of group sparsity~\cite{murdock2015blockout,graham2014spatially}.

If a foreground image is reconstructed by feature maps that are crosswise sparse, as defined by Eq.~\ref{eq:spartial_sparsity}, the foreground image is essentially reconstructed by a few vectors in the feature maps. As a result, those vectors must represent salient objects in the foreground image- nuclei, since the CAE aims to minimize the reconstruction error.

\subsubsection{Ensuring Crosswise Sparsity}
Crosswise sparsity defined by Eq.~\ref{eq:spartial_sparsity} is not achievable using conventional sparsification methods~\cite{ng2011sparse} that can only satisfy Eq.~\ref{eq:sparsity}. We introduce a binary matrix $D$ with its $i,j$-th entry $D^{i,j}$ indicating if $X_l^{i,j}$ are activated for any $l$ or not:
\begin{equation}
D^{i,j} = \mathbbm{1}\Big(\sum_l \mathbbm{1}(X_l^{i,j} \ne 0) \, > \, 0\Big)\text{.} 
\label{eq:detection_map}
\end{equation} Therefore Eq.~\ref{eq:spartial_sparsity} becomes:
\begin{equation}
\dfrac{\sum_{i,j} D^{i,j}}{s^2} \ll 1 \text{.}
\label{eq:redefined_traverse_sparsity}
\end{equation}
The foreground feature maps $X_1$, $X_2$, \dots $X_f$ are crosswise sparse, {\em iff} there exists a matrix $D$ that satisfies Eq.~\ref{eq:detection_map} and Eq.~\ref{eq:redefined_traverse_sparsity}. To satisfy Eq.~\ref{eq:detection_map}, we design the CAE to generate a binary sparse feature map that represents $D$. The CAE computes $X_l$ based on a dense feature map $X_l'$ and $D$ by element-wise multiplication:
\begin{equation}
X_l = X_l' \odot D\text{.}
\label{eq:element-wise-multi}
\end{equation}
We call the feature map $D$ the detection map, shown in Fig.~\ref{fig:pipeline}. The dense feature map $X_l'$ is automatically learned by the CAE by minimizing the reconstruction error.

The proposed CAE also computes the $D$ that satisfies Eq.~\ref{eq:redefined_traverse_sparsity}. Notice that Eq.~\ref{eq:redefined_traverse_sparsity} is equivalent to the conventional sparsity defined by Eq.~\ref{eq:sparsity}, when the total number of feature maps $f=1$ and $X_f$ is a binary feature map. Therefore, Eq.~\ref{eq:redefined_traverse_sparsity} can be satisfied by existing sparsification methods. A standard sparsification methods is to add a sparsity penalty term in the loss function~\cite{ng2011sparse}. This method requires parameter tuning to achieve the desired expected sparsity. The $k$-sparse method~\cite{makhzani2013k} guarantees that exactly $k$ neurons will be activated in $D$, where $k$ is a predefined constant. However, in tissue images, the number of nuclei per image varies; the sparsity rate also should vary.

In this paper, we propose to use a threshold based method that guarantees an overall expected predefined sparsity rate, even though the sparsity rate for each CAE's input can vary. We compute the binary sparse feature map $D$ as output from an automatically learned input dense feature map $D'$:
\begin{equation}
D^{i,j} = \mathrm{sig}\big(r(D'^{i,j}-t)\big)\text{,}
\label{eq:sparsity-by-sigmoid}
\end{equation}where $\mathrm{sig}(\cdot)$ is the sigmoid function, $r$ is a predefined slope, and $t$ is an automatically computed threshold. We choose $r=20$ in all experiments, making $D$ a binary matrix in practice. Different $r$ values do not affect the performance significantly based on our experience. Our CAE computes a large $t$ in the training phase, which results in a sparse $D$. We define the expected sparsity rate as $p\%$, which can be set according to the average number of nuclei per image (we use $p=1.6$ in all experiments), we compute $t$ as:
\begin{equation}
t = \mathrm{E}[\mathrm{percentile}_p(D'^{i,j})] \text{,}
\label{eq:threshold}
\end{equation}where $\mathrm{percentile}_p(D'^{i,j})$ is the $p$-th percentile of $D'^{i,j}$ for all $i,j$, given a particular CAE's input image. We compute $t$ using the running average method:
\begin{equation}
t \leftarrow (1-\alpha) t + \alpha \,\mathrm{percentile}_p(D'^{i,j}) \text{.}
\label{eq:running_average}
\end{equation} We set the constant $\alpha=0.1$ in all experiments. This running average approach is also used by batch normalization~\cite{ioffe2015batch}. To make sure the running average of $t$ converges, we also use batch normalization on $D'^{i,j}$ to normalize the distribution of $D'^{i,j}$ in each stochastic gradient descent batch. In total, three parameters are introduced in our CAE: $r$, $p$, and $\alpha$. The sparsity rate $p$ can be decided based on the dataset easily. The other two parameters do not affect the performance significantly in our experiments.

With crosswise sparsity, each vector in the foreground feature maps can possibly encode multiple nuclei. To achieve one-on-one correspondence between nuclei and encoded vectors, we simply reduce the size of the encoding neurons' receptive fields, such that a vector encodes a small region the size of which is close to the size of a nucleus.

\section{Experiments}
\label{sec:exp}
We initialize CNNs with our crosswise sparse CAEs. We empirically evaluate this approach on four datasets: a self-collected lymphocyte-rich region classification dataset, a self-collected individual lymphocyte classification dataset, the nuclear shape and attribute classification dataset,
and the MICCAI 2015 nucleus segmentation challenge dataset~\cite{miccai2015segmentation}. The results show that the proposed method achieves better results than other methods. 

\begin{figure*}[t]
 \subcaptionbox*{}{\rotatebox[origin=t]{90}{Image}}
 \includegraphics[width=0.118\linewidth]{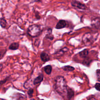}
 \includegraphics[width=0.118\linewidth]{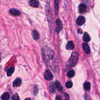}
 \includegraphics[width=0.118\linewidth]{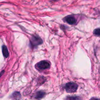}
 \includegraphics[width=0.118\linewidth]{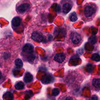}
 \includegraphics[width=0.118\linewidth]{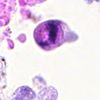}
 \includegraphics[width=0.118\linewidth]{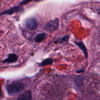}
 \includegraphics[width=0.118\linewidth]{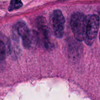}
 \includegraphics[width=0.118\linewidth]{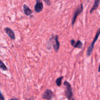}
 \vspace{-0.5cm}\\

 \subcaptionbox*{}{\rotatebox[origin=t]{90}{Detection}}
 \hspace{0.02cm}
 \includegraphics[width=0.118\linewidth]{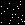}
 \includegraphics[width=0.118\linewidth]{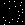}
 \includegraphics[width=0.118\linewidth]{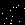}
 \includegraphics[width=0.118\linewidth]{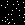}
 \includegraphics[width=0.118\linewidth]{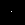}
 \includegraphics[width=0.118\linewidth]{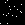}
 \includegraphics[width=0.118\linewidth]{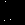}
 \includegraphics[width=0.118\linewidth]{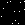}
 \vspace{-0.5cm}\\

 \subcaptionbox*{}{\rotatebox[origin=t]{90}{Foreground}}
 \includegraphics[width=0.118\linewidth]{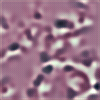}
 \includegraphics[width=0.118\linewidth]{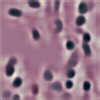}
 \includegraphics[width=0.118\linewidth]{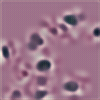}
 \includegraphics[width=0.118\linewidth]{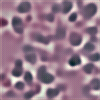}
 \includegraphics[width=0.118\linewidth]{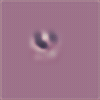}
 \includegraphics[width=0.118\linewidth]{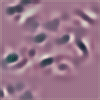}
 \includegraphics[width=0.118\linewidth]{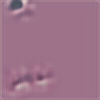}
 \includegraphics[width=0.118\linewidth]{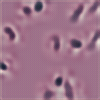}
 \vspace{-0.5cm}\\

 \subcaptionbox*{}{\rotatebox[origin=t]{90}{Background}}
 \includegraphics[width=0.118\linewidth]{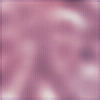}
 \includegraphics[width=0.118\linewidth]{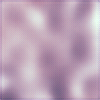}
 \includegraphics[width=0.118\linewidth]{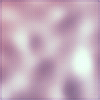}
 \includegraphics[width=0.118\linewidth]{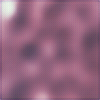}
 \includegraphics[width=0.118\linewidth]{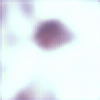}
 \includegraphics[width=0.118\linewidth]{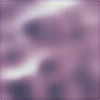}
 \includegraphics[width=0.118\linewidth]{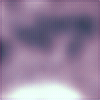}
 \includegraphics[width=0.118\linewidth]{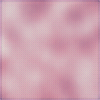}
 \vspace{-0.5cm}\\

 \centering\subcaptionbox*{}{\rotatebox[origin=t]{90}{Reconstruct}}
 \hspace{0.001cm}
 \includegraphics[width=0.1175\linewidth]{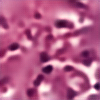}
 \includegraphics[width=0.1175\linewidth]{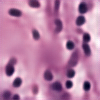}
 \includegraphics[width=0.1175\linewidth]{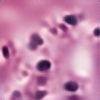}
 \includegraphics[width=0.1175\linewidth]{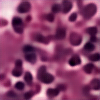}
 \includegraphics[width=0.1175\linewidth]{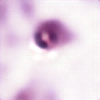}
 \includegraphics[width=0.1175\linewidth]{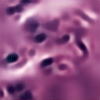}
 \includegraphics[width=0.1175\linewidth]{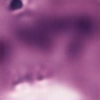}
 \includegraphics[width=0.1175\linewidth]{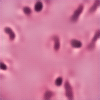}
 \vspace{-0.5cm}

   \caption{Randomly selected examples of unsupervised nucleus detection and foreground, background image representation (reconstruction) results. We show the detection map in Fig.~\ref{fig:pipeline} in the second row. The reconstructed image (last row) is the pixel-wise summation of the reconstructed foreground and background images. We can see that our CAE can decompose input images to foreground and background, and detect and represent (reconstruct) nuclei in the foreground.}
\label{fig:cae-examples}
\end{figure*}

\begin{table}[h!]
	\centering
	\begin{tabular}{c | c | c | c | l}
    \hline
	 Part & Layer & Kernel size & Stride & Output size \\
    \hline
	\hline
     \multirow{9}{*}{1} & Input & - & - & $100^2\times 3$ \\
     & Conv & 5$\times$5 & 1 & $100^2\times 100$ \\
     & Conv & 5$\times$5 & 1 & $100^2\times 120$ \\
     & Pool & 2$\times$2 & 2 & $50^2\times 120$ \\
     & Conv & 3$\times$3 & 1 & $50^2\times 240$ \\
     & Conv & 3$\times$3 & 1 & $50^2\times 320$ \\
     & Pool & 2$\times$2 & 2 & $25^2\times 320$ \\
     & Conv & 3$\times$3 & 1 & $25^2\times 640$ \\
     & Conv & 3$\times$3 & 1 & $25^2\times 1024$ \\
    \hline
     \multirow{2}{*}{2} & Conv & 1$\times$1 & 1 & $25^2\times 100$ \\
     & Conv & 1$\times$1 & 1 & $25^2\times 1$ \\
    \hline
     \multirow{2}{*}{3} & Conv & 1$\times$1 & 1 & $25^2\times 640$ \\
     & Conv & 1$\times$1 & 1 & $25^2\times 100$ \\
    \hline
     \multirow{4}{*}{4} & Conv & 1$\times$1 & 1 & $25^2\times 128$ \\
     & Pool & 5$\times$5 & 5 & $5^2\times 128$ \\
     & Conv & 3$\times$3 & 1 & $5^2\times 64$ \\
     & Conv & 1$\times$1 & 1 & $5^2\times 5$ \\
    \hline
     \multirow{1}{*}{5} & Thres. & \multicolumn{2}{|c|}{Defined by Eq.~\ref{eq:sparsity-by-sigmoid}} & $25^2\times 1$ \\
    \hline
     \multirow{1}{*}{6} & Filter. & \multicolumn{2}{|c|}{Defined by Eq.~\ref{eq:element-wise-multi}} & $25^2\times 100$ \\
    \hline
     \multirow{9}{*}{7} & Deconv & 3$\times$3 & 1 & $25^2 \times 1024$ \\
     & Deconv & 3$\times$3 & 1 & $25^2 \times 640$ \\
     & Deconv & 4$\times$4 & 0.5 & $50^2 \times 640$ \\
     & Deconv & 3$\times$3 & 1 & $50^2 \times 320$ \\
     & Deconv & 3$\times$3 & 1 & $50^2 \times 320$ \\
     & Deconv & 4$\times$4 & 0.5 & $100^2 \times 320$ \\
     & Deconv & 5$\times$5 & 1 & $100^2 \times 120$ \\
     & Deconv & 5$\times$5 & 1 & $100^2 \times 100$ \\
     & Deconv & 1$\times$1 & 1 & $100^2 \times 3$ \\
	\hline
     \multirow{12}{*}{8} & Deconv & 3$\times$3 & 1 & $5^2 \times 256$ \\
     & Deconv & 3$\times$3 & 1 & $5^2 \times 128$ \\
     & Deconv & 9$\times$9 & 0.2 & $25^2 \times 128$ \\
     & Deconv & 3$\times$3 & 1 & $25^2 \times 128$ \\
     & Deconv & 3$\times$3 & 1 & $25^2 \times 128$ \\
     & Deconv & 4$\times$4 & 0.5 & $50^2 \times 128$ \\
     & Deconv & 3$\times$3 & 1 & $50^2 \times 64$ \\
     & Deconv & 3$\times$3 & 1 & $50^2 \times 64$ \\
     & Deconv & 4$\times$4 & 0.5 & $100^2 \times 64$ \\
     & Deconv & 5$\times$5 & 1 & $100^2 \times 32$ \\
     & Deconv & 5$\times$5 & 1 & $100^2 \times 32$ \\
     & Deconv & 1$\times$1 & 1 & $100^2 \times 3$ \\
	\hline
	\end{tabular}
\caption{Layer setup of different parts in our CAE. Please refer to Fig.~\ref{fig:pipeline} for the overall network architecture.}
\label{tab:net_layers}
\end{table}

\subsection{Datasets} \label{sec:datasets} 

{\em Dataset for Unsupervised Learning.}
We collected 0.5 million unlabeled small images randomly cropped from 400 lung adenocarcinoma histopathology images obtained from the public TCGA repository~\cite{TCGAdataset}. The cropped images are 100$\times$100 pixels in 20X (0.5 microns per pixel). We will refer to cropped images simply as images in the rest of this section.

\begin{figure}[t]
\begin{center}
   \includegraphics[width=0.32\linewidth]{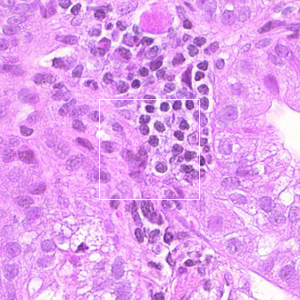}
   \includegraphics[width=0.32\linewidth]{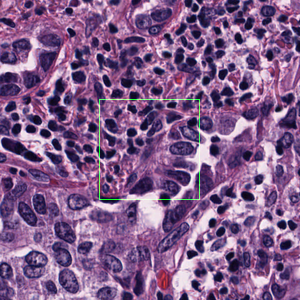}
   \includegraphics[width=0.32\linewidth]{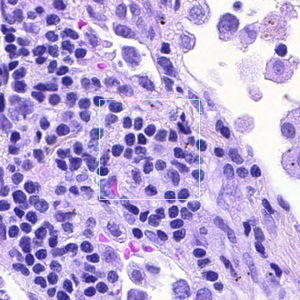}\vspace{0.1cm}
   \includegraphics[width=0.32\linewidth]{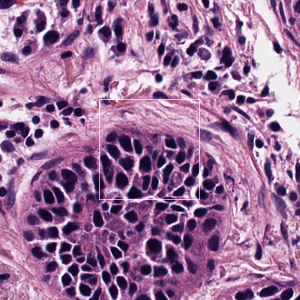}
   \includegraphics[width=0.32\linewidth]{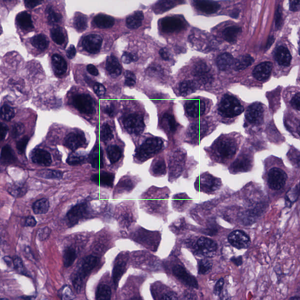}
   \includegraphics[width=0.32\linewidth]{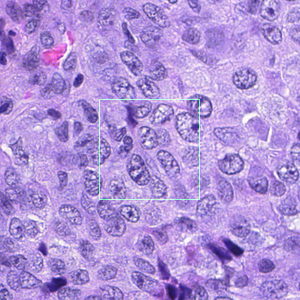}
\end{center}
\vspace{-0.25cm}
   \caption{Examples of the lymphocyte rich region dataset. Top: lymphocyte rich images. Bottom: non-lymphocyte-rich images. A pathologist labeled 23,356 images depending on the percentage of lymphocytes in the center $100\times 100$ pixels of the image (framed in green). The peripheral pixels provide context information to the pathologist and automatic classification models.}
\label{fig:lym_training}
\end{figure}

{\em Dataset for Lymphocyte Classification Experiments (Sec~\ref{sec:lym-app}).} Lymphocytes and plasma cells are types of white blood cells in the immune system. Automatic recognition of lymphocytes and plasma cells is very important in many situations including the study of cancer immunotherapy~\cite{galon2006type,salgado2015evaluation,turkki2016antibody}. We collected a dataset of 23,356 images labeled as lymphocyte (including plasma cells) rich or not. These images are cropped from lung adenocarcinoma tissue images in the publicly available TCGA dataset~\cite{TCGAdataset}. Each image is 300$\times$300 pixels in $20X$ (0.5 microns per pixel). A pathologist labeled these images, according to the percentage of lymphocytes in the center 100$\times$100 pixels of the image. The peripheral pixels provide context information to the pathologist and to automatic classification models. Overall, around 6\% of the images are labeled as lymphocyte rich. We show examples of the training set in Fig.~\ref{fig:lym_training}.

{\em Dataset for Classifying Individual Lymphocytes Experiments (Sec.~\ref{sec:lym-individual-app}).}
We collected a dataset of 1785 images of individual objects that were labeled lymphocyte or non-lymphocyte by a pathologist. These 1785 images were cropped from 12 representative lung adenocarcinoma whole slide tissue images from the TCGA repository. We use labeled images cropped from 10 whole slide tissue images as the training data and the rest as the test dataset.

{\em Dataset for Nuclear Shape and Attribute Classification Experiments (Sec.~\ref{sec:nuclear-shape-attribute}).}
We apply our method on an existing dataset~\cite{murthy2017center} for nuclear shape and attribute classification. The dataset consists of 2000 images of nuclei labeled with fifteen morphology attributes and shapes.

{\em Dataset for Nucleus Segmentation Experiments (Sec.~\ref{sec:nucleus-app}).} We test our method for nucleus segmentation using the MICCAI 2015 nucleus segmentation challenge dataset~\cite{miccai2015segmentation} which contains 15 training images and 18 testing images with a typical resolution of 500$\times$500. The ground truth masks of nuclei are provided in the training dataset.

\subsection{CAE Architecture}
\label{sec:network_archi}
The CAEs in all three classification experiments (Sec.~\ref{sec:lym-app}, Sec.~\ref{sec:lym-individual-app} and Sec.~\ref{sec:nuclear-shape-attribute}) are trained on the unlabeled dataset with the same architecture illustrated in Fig.~\ref{fig:pipeline} and Tab.~\ref{tab:net_layers}. Note that we apply batch normalization~\cite{ioffe2015batch} before the leaky ReLU activation functions~\cite{maas2013rectifier} in all layers.

The average nucleus size in the dataset for nucleus segmentation experiments (Sec.~\ref{sec:nucleus-app}) is around $20\times 20$ pixels. Therefore pooling layers can discard important spatial information which is important for pixel-wise segmentation. The U-net~\cite{ronneberger2015u} addresses this issue by adding skip connections. However, we find in practice that eliminating pooling layers completely yields better performance. The computation complexity is very high for a network without any pooling layers. Thus, compared to Tab.~\ref{tab:net_layers}, we use smaller input dimensions ($40 \times 40$) and fewer (80 to 200) feature maps in the CAE. Other settings of the CAE for segmentation remain unchanged.

\subsection{CNN Architecture}
\label{sec:network_archi_cnn}
We construct supervised CNNs based on trained CAEs. For classification tasks, the supervised CNN contain Parts 1-6 of the CAE. We initialize the parameters in these layers to be the same as the parameters in the CAE. We attach four $1\times 1$ convolutional layers after the foreground encoding layer and two $3\times 3$ convolutional layers after the background encoding layer. Each added layer has 320 convolutional filters. We then apply global average pooling on the two branches. The pooled features are then concatenated together, followed by a final classification layer with sigmoid activation function. For the segmentation task, the supervised CNN only contains Parts 1, 3 of the CAE. We attach six $3\times 3$ convolutional layers followed by a segmentation layer. The segmentation layer is the same to the patch-CNN's~\cite{vicente2016large} segmentation layer which is a fully-connected layer with sigmoid activation function followed by reshaping. For all tasks, we randomly initialize the parameters of these additional layers. We train the parameters of the added layers until convergence before fine-tuning the whole network.

\subsection{Learning Details}
We train our CAE on the unlabeled dataset, 
minimizing the pixel-wise root mean squared error between the input images and the reconstructed images. We use stochastic gradient descent with batch size 32, learning rate $0.03$ and momentum $0.9$. The loss converges after 6 epochs. We show randomly selected examples of the nucleus detection feature map as well as the reconstructed foreground and background images in Fig.~\ref{fig:cae-examples}.

For the CNN (constructed from the CAE) training, we use stochastic gradient descent with batch size, learning rate, and momentum selected for each task independently. For all tasks, we divide the learning rate by 10 when the error has plateaued. We use sigmoid as the nonlinearity function in the last layer and log-likelihood as the loss function. We apply three types of data augmentation. First, the input images are randomly cropped from a larger image. Second, the colors of the input images are randomly perturbed. Third, we randomly rotate and mirror the input images. During testing, we average the predictions of 25 image crops. We implemented our CAE and CNN using Theano~\cite{2016arXiv160502688short}. We trained the CAE and CNN on a single Tesla K40 GPU.

\subsection{Methods Tested}
We describe our method (abbreviated as CSP-CNN) and other tested methods below:
\begin{description}[leftmargin=0.2in]
\setlength\itemsep{0.18em}
\item[CSP-CNN] CNN initialized by our proposed crosswise sparse CAE shown in Fig.~\ref{fig:pipeline}. The exact CNN construction is described in Sec.~\ref{sec:network_archi_cnn}. We set the sparsity rate to $1.6\%$, such that the number of activated foreground feature map locations roughly equals to the average number of nuclei per image in the unsupervised training set.
\item[SUP-CNN] A fully supervised CNN. Its architecture is similar to our CSP-CNN except that: 1). There is no background representation branch (no Part 4, 8 in Fig.~\ref{fig:pipeline}). 2). There is no nucleus detection branch (no Part 2, 5 in Fig.~\ref{fig:pipeline}). The SUP-CNN has a very standard architecture, at the same time similar to our CSP-CNN.
\item[U-NET] We use the authors' U-net architecture and implementation~\cite{ronneberger2015u} for nucleus segmentation. We test five U-nets with the same architecture but different number of feature maps per layer and select the best performing network. All five U-nets perform similarly.
\item[DEN-CNN] CNN initialized by a conventional Convolutional Autoencoder (CAE) without the sparsity constraint. Its architecture is similar to our CSP-CNN except that it there is no nucleus detection branch. In particular, there is no Part 2 and Part 5 in Fig.~\ref{fig:pipeline} and Part 6 is an identity mapping layer.
\item[SP-CNN] CNN initialized by a sparse CAE without the crosswise constraint. Its architecture is similar to our CSP-CNN except that it has no nucleus detection branch and uses the conventional sparsity constraint defined by Eq.~\ref{eq:sparsity}. In particular, there is no Part 2 and Part 5 in Fig.~\ref{fig:pipeline} and Part 6 is a thresholding layer: define its input as $D'$, its output $D=\mathrm{ReLU}\big(D'-t\big)$, where $t$ is obtained in the same way defined by Eq.~\ref{eq:threshold}. We set the sparsity rate to $1.6\%$ which equals to the rate we use in CSP-CNN.
\item[VGG16] We finetune the VGG 16-layer network~\cite{simonyan2014very} which is pretrained on ImageNet~\cite{russakovsky2015imagenet}. Fine-tuning the VGG16 network has been shown to be robust for pathology image classification~\cite{xu2015deep,hou2016automatic}.
\end{description}

\begin{table}[h!]
	\centering
	\begin{tabular}{l | c | c | c}
	\hline
    \multirow{3}{*}{Methods} & \multicolumn{3}{c}{Datasets} \\
    \cline{2-4}
     & Lym- & Individual & Nuclear Attr\\
     & region & Lym & \&Shape~\cite{murthy2017center}\\
    \hline
    \hline
     SUP-CNN & 0.6985 & 0.4936 & 0.8487\\
     DEN-CNN & 0.8764 & 0.5576 & 0.8656\\
     SP-CNN & 0.9188 & 0.6262 & 0.8737\\
     CSP-CNN & \textbf{0.9526} & \textbf{0.7856} & \textbf{0.8788}\\
     CSP-CNN & \multirow{2}{*}{0.9215} & \multirow{2}{*}{0.7135} & \multirow{2}{*}{0.7128}\\
     (5\% data) &  &\\
    \hline
     Unsupervised & \multirow{2}{*}{-} & \multirow{2}{*}{0.7132} & \multirow{2}{*}{-}\\
     features \cite{Naiyun2017evaluation} & & \\
     Semi-supervised & \multirow{2}{*}{-} & \multirow{2}{*}{-} & \multirow{2}{*}{0.8570}\\
     CNN \cite{murthy2017center} & & \\
     VGG16 \cite{simonyan2014very} & 0.9176 & 0.6925 & 0.8480 \\
     
	\hline
	\end{tabular}
\caption{Classification results measured by AUROC on there tasks described in Sec.~\ref{sec:lym-app}, Sec.~\ref{sec:lym-individual-app}, Sec.~\ref{sec:nuclear-shape-attribute}. The proposed CSP-CNN outperforms the other methods significantly. Comparing the results of SP-CNN and our CSP-CNN, we can see that the proposed crosswise constraint boosts performance significantly. Even with only 5\% labeled training data, our CSP-CNN (5\% data) outperforms other methods on the first two datasets. The CSP-CNN (5\% data) fails on the third dataset because when only using 5\% of the training data, 5 out of 15 classes have less than 2 positive training instances which are too few for CNN training.}
\label{tab:lym-results}
\end{table}

\subsection{Classifying Lymphocyte-rich Regions}
\label{sec:lym-app}
We use 20,876 images as the training set and the remaining 2,480 images as testing set. We use the Area Under ROC Curve (AUROC) as the evaluation metric. The results are shown in Tab.~\ref{tab:lym-results}. Our proposed CSP-CNN achieves the best result on this dataset. Our CSP-CNN reduces the error of the best performing baseline SP-CNN by \textbf{42\%}. Furthermore, with only 5\% of the training data, our CSP-CNN (5\% data) outperforms SP-CNN. The only difference between CSP-CNN and SP-CNN is the crosswise constraint, with which our CAE is capable of unsupervised nucleus detection. This supports our claim that our crosswise sparsity is essential to high performance.

\subsection{Classifying Individual Lymphocytes}
\label{sec:lym-individual-app}
We compare our method with an existing unsupervised nucleus detection and feature extraction method \cite{Naiyun2017evaluation}.  We split training and test images 4 times and average the results. As the baseline method we carefully tuned a recently published unsupervised nucleus detection and feature extraction method \cite{Naiyun2017evaluation}, which is based on level sets, and applied a multi-layer neural network on top of the extracted features. We should note that the feature extraction step and the classification step have to be tuned separately in the baseline method, whereas our CSP-CNN method (Sec.~\ref{sec:lym-app}) can be trained end-to-end. We show results in Tab. \ref{tab:lym-results}. Our CSP-CNN reduced the error of the SP-CNN by \textbf{25\%}.

\subsection{Nuclear Shape and Attribute Classification}
\label{sec:nuclear-shape-attribute}
We adopt the same 5-fold training and testing data separation protocol and report the results in Tab. \ref{tab:lym-results}. On this dataset the improvement of our method over the state-of-the-art is less significant than the improvement on other datasets because the images of nuclei are results of a fixed nucleus detection method which we cannot fine-tune with our proposed method.

\begin{figure}[t]
\begin{center}
   \includegraphics[trim=4cm 4cm 4cm 5cm,clip, width=0.32\linewidth]{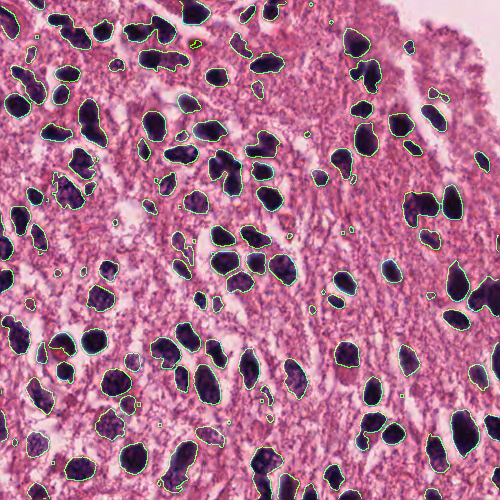}
   \includegraphics[trim=3cm 2cm 3cm 2cm,clip, width=0.32\linewidth]{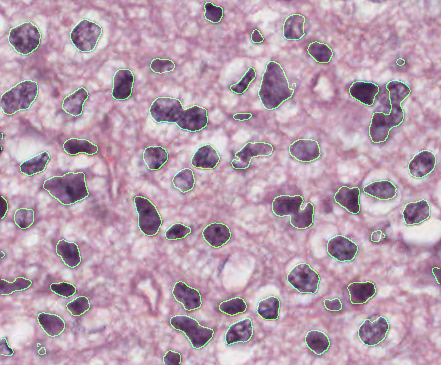}
   \includegraphics[trim=3cm 3cm 2cm 3cm,clip, width=0.32\linewidth]{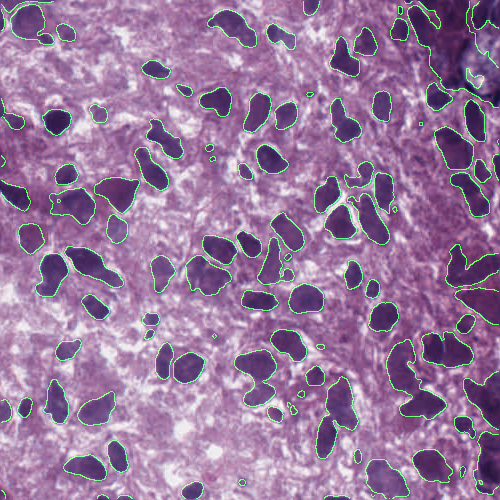}
\end{center}
\vspace{-0.25cm}
   \caption{Randomly selected examples of nucleus segmentation using our CSP-CNN, on the MICCAI 2015 nucleus segmentation challenge dataset (best viewed in color). The segmentation boundaries are displayed in green.}
\label{fig:seg_examples}
\end{figure}

\subsection{Nucleus Segmentation}
\label{sec:nucleus-app}
We use a sliding window approach to train and test our CNNs. A CNN outputs a feature map of the same size as its input. For evaluation, we followed the standard metric used in the MICCAI challenge: the DICE-average (average of two different versions of the DICE coefficient). We show results in Tab.~\ref{tab:segmentation-results}. The proposed method achieves a significantly higher score than that of the challenge winner~\cite{chen2017dcan} and U-net~\cite{ronneberger2015u}. Because the size of nuclei are only around $20 \times 20$ pixels, we eliminate our network's pooling layers completely to preserve spatial information. We believe this is an important reason our method outperforms U-net. We show randomly selected segmentation examples in Fig.~\ref{fig:seg_examples}.
\begin{table}[h!]
	\centering
	\begin{tabular}{l | l}
	\hline
	 Methods & DICE-average \\
    \hline
    \hline
     SUP-CNN & 0.8216 \\
     DEN-CNN & 0.8235 \\
     SP-CNN & 0.8338 \\
     CSP-CNN & \textbf{0.8362} \\
     CSP-CNN (5\% data) & 0.8205 \\
	\hline
     Challenge winner~\cite{chen2017dcan} & 0.80 \\
     U-net~\cite{ronneberger2015u} & 0.7942 \\
	\hline
	\end{tabular}
\caption{Nucleus segmentation results on the MICCAI 2015 nucleus segmentation challenge dataset. Our CSP-CNN outperforms the highest challenge score which is a DICE-average of 0.80, even with only 5\% of the sliding windows during training. On this dataset we do not use pooling layers in the networks, because we find that pooling layers discard important spatial information, since the size of nuclei are only around $20 \times 20$ pixels.}
\label{tab:segmentation-results}
\end{table}

\section{Conclusions}
\label{sec:conclusions}
We propose a crosswise sparse Convolutional Autoencoder (CAE) that for the first time, is capable of unsupervised nucleus detection and feature extraction simultaneously. We advocate that this CAE should be used to initialize classification or segmentation Convolutional Neural Networks (CNN) for supervised training. In this manner, the nucleus detection, feature extraction, and classification or segmentation steps are trained end-to-end. Our experimental results with four challenging datasets show that our proposed crosswise sparsity is essential to state-of-the-art results. In a future work we plan to perform unsupervised nucleus segmentation with the proposed CAE.

{\small
\bibliographystyle{ieee}
\bibliography{ref}
}

\end{document}